# Using Deep Learning for Image-Based Plant Disease Detection

Sharada Prasanna Mohanty[1,2], David Hughes[3,4,5], and Marcel Salathé[1,2,6]

[1]Digital Epidemiology Lab, EPFL, Switzerland; [2]School of Life Sciences, EPFL, Switzerland; [3]Department of Entomology, College of Agricultural Sciences, Penn State University, USA; [4]Department of Biology, Eberly College of Sciences, Penn State University, USA; [5]Center for Infectious Disease Dynamics, Huck Institutes of Life Sciences, Penn State University, USA; [6]School of Computer and Communication Sciences, EPFL, Switzerland

This manuscript was compiled on April 15, 2016

**Crop diseases are a major threat to food security, but their rapid identification remains difficult in many parts of the world due to the lack of the necessary infrastructure. The combination of increasing global smartphone penetration and recent advances in computer vision made possible by deep learning has paved the way for smartphone-assisted disease diagnosis. Using a public dataset of 54,306 images of diseased and healthy plant leaves collected under controlled conditions, we train a deep convolutional neural network to identify 14 crop species and 26 diseases (or absence thereof). The trained model achieves an accuracy of 99.35% on a held-out test set, demonstrating the feasibility of this approach. When testing the model on a set of images collected from trusted online sources - i.e. taken under conditions different from the images used for training - the model still achieves an accuracy of 31.4%. While this accuracy is much higher than the one based on random selection (2.6%), a more diverse set of training data is needed to improve the general accuracy. Overall, the approach of training deep learning models on increasingly large and publicly available image datasets presents a clear path towards smartphone-assisted crop disease diagnosis on a massive global scale.**

Deep Learning | Crop Diseases | Digital Epidemiology

**M**odern technologies have given human society the ability to produce enough food to meet the demand of more than 7 billion people. However, food security remains threatened by a number of factors including climate change[1], the decline in pollinators[2], plant diseases [3], and others. Plant diseases are not only a threat to food security at the global scale, but can also have disastrous consequences for smallholder farmers whose livelihoods depend on healthy crops. In the developing world, more than 80 percent of the agricultural production is generated by smallholder farmers [4], and reports of yield loss of more than 50% due to pests and diseases are common [5]. Furthermore, the largest fraction of hungry people (50%) live in smallholder farming households [6], making smallholder farmers a group that's particularly vulnerable to pathogen-derived disruptions in food supply.

Various efforts have been developed to prevent crop loss due to diseases. Historical approaches of widespread application of pesticides have in the past decade increasingly been supplemented by integrated pest management (IPM) approaches [7]. Independent of the approach, identifying a disease correctly when it first appears is a crucial step for efficient disease management. Historically, disease identification has been supported by agricultural extension organizations or other institutions such as local plant clinics. In more recent times, such efforts have additionally been supported by providing information for disease diagnosis online, leveraging the increasing internet penetration worldwide. Even more recently, tools based on mobile phones have proliferated, taking advantage of the historically unparalleled rapid uptake of mobile phone technology in all parts of the world[8].

Smartphones in particular offer very novel approaches to help identify diseases because of their tremendous computing power, high-resolution displays, and extensive built-in sets of accessories such as advanced HD cameras. It is widely estimated that there will be between 5 and 6 billion smartphones on the globe by 2020. At the end of 2015, already 69% of the world's population had access to mobile broadband coverage, and mobile broadband penetration reached 47% in 2015, a 12-fold increase since 2007[8]. The combined factors of widespread smartphone penetration, HD cameras, and high performance processors in mobile devices lead to a situation where disease diagnosis based on automated image recognition, if technically feasible, can be made available at an unprecedented scale. Here, we demonstrate the technical feasibility using a deep learning approach utilizing 54,306 images of 14 crop species with 26 diseases (or healthy) made openly available through the project PlantVillage[9]. An example of each crop - disease pair cann bee seen in Figure 1.

Computer vision, and object recognition in particular, has made tremendous advances in the past few years. The PASCAL VOC Challenge[10], and more recently the Large Scale Visual Recognition Challenge (ILSVRC)[11] based on the ImageNet dataset[12] have been widely used as benchmarks for numerous visualization-related problems in computer vision, including object classification. In 2012, a large, deep convolutional neural network achieved a top-5 error of 16.4% for the classification of images into 1,000 possible categories[13]. In the following three years, various advances in deep convolutional neural networks lowered the error rate to 3.57% [13] [14] [15] [16] [17]. While training large neural networks can be very time-consuming, the trained models can classify images

> **Significance Statement**
>
> Crop diseases remain a major threat to food supply worldwide. This paper demonstrates the technical feasibility of a deep learning approach to enable automatic disease diagnosis through image recognition. Using a public dataset of 54,306 images of diseased and healthy plant leaves, a deep convolutional neural network is trained to classify crop species and disease status of 38 different classes containing 14 crop species and 26 diseases, achieving an accuracy of over 99%.

Corresponding author: marcel.salathe@epfl.ch

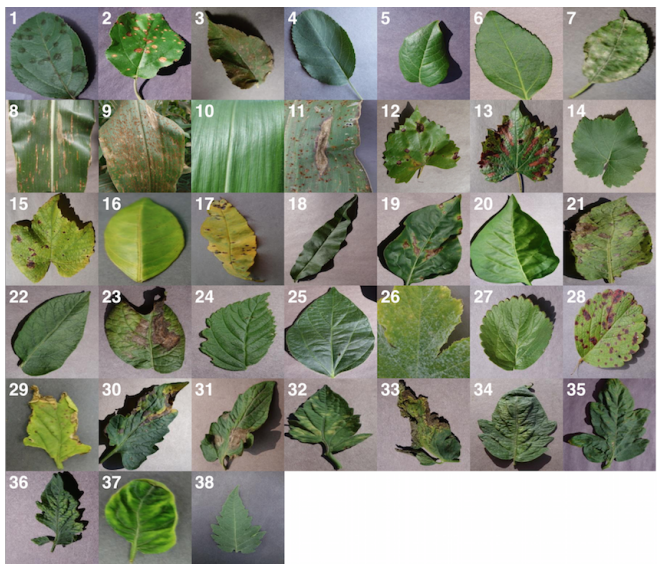

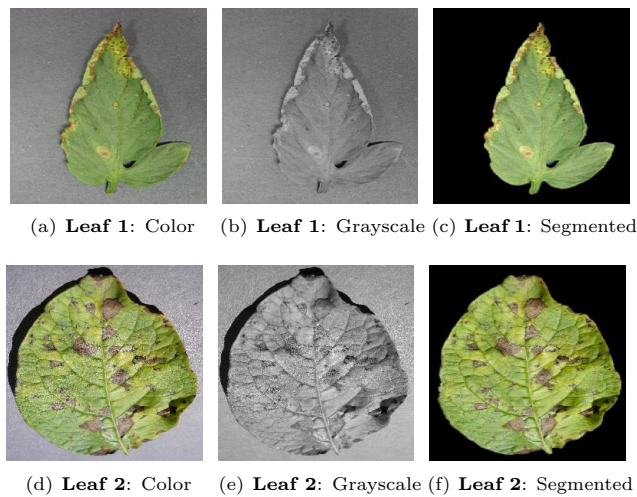

**Fig. 1.** Example of leaf images from the PlantVillage dataset, representing every crop-disease pair used. 1) Apple Scab, *Venturia inaequalis* 2) Apple Black Rot, *Botryosphaeria obtusa* 3) Apple Cedar Rust, *Gymnosporangium juniperi-virginianae* 4) Apple healthy 5) Blueberry healthy 6) Cherry healthy 7) Cherry Powdery Mildew, *Podosphaera spp.* 8) Corn Gray Leaf Spot, *Cercospora zeae-maydis* 9) Corn Common Rust, *Puccinia sorghi* 10) Corn healthy 11) Corn Northern Leaf Blight, *Exserohilum turcicum* 12) Grape Black Rot, *Guignardia bidwellii*, 13) Grape Black Measles (Esca), *Phaeomoniella aleophilum, Phaeomoniella chlamydospora* 14) Grape Healthy 15) Grape Leaf Blight, *Pseudocercospora vitis* 16) Orange Huanglongbing (Citrus Greening), *Candidatus Liberibacter spp.* 17) Peach Bacterial Spot, *Xanthomonas campestris* 18) Peach healthy 19) Bell Pepper Bacterial Spot, *Xanthomonas campestris* 20) Bell Pepper healthy 21) Potato Early Blight, *Alternaria solani* 22) Potato healthy 23) Potato Late Blight, *Phytophthora infestans* 24) Raspberry healthy 25) Soybean healthy 26) Squash Powdery Mildew, *Erysiphe cichoracearum, Sphaerotheca fuliginea* 27) Strawberry Healthy 28) Strawberry Leaf Scorch, *Diplocarpon earlianum* 29) Tomato Bacterial Spot, *Xanthomonas campestris pv. vesicatoria* 30) Tomato Early Blight, *Alternaria solani* 31) Tomato Late Blight, *Phytophthora infestans* 32) Tomato Leaf Mold, *Fulvia fulva* 33) Tomato Septoria Leaf Spot, *Septoria lycopersici* 34) Tomato Two Spotted Spider Mite, *Tetranychus urticae* 35) Tomato Target Spot, *Corynespora cassiicola* 36) Tomato Mosaic Virus 37) Tomato Yellow Leaf Curl Virus 38) Tomato healthy

**Fig. 2.** Sample images from the three different versions of the PlantVillage dataset used in various experimental configurations.

very quickly, which makes them also suitable for consumer applications on smartphones.

In order to develop accurate image classifiers for the purposes of plant disease diagnosis, we needed a large, verified dataset of images of diseased and healthy plants. Until very recently, such a dataset did not exist, and even smaller datasets were not freely available. To address this problem, the PlantVillage project has begun collecting tens of thousands of images of healthy and diseased crop plants [9], and has made them openly and freely available. Here, we report on the classification of 26 diseases in 14 crop species using 54,306 images with a convolutional neural network approach. We measure the performance of our models based on their ability to predict the correct crop-diseases pair, given 38 possible classes. The best performing model achieves a mean $F_1$ score of 0.9934 (overall accuracy of 99.35%), hence demonstrating the technical feasibility of our approach. Our results are a first step towards a smartphone-assisted plant disease diagnosis system.

### Results

At the outset, we note that on a dataset with 38 class labels, random guessing will only achieve an overall accuracy of 2.63% on average. Across all our experimental configurations, which include three visual representations of the image data (see Figure 2), the overall accuracy we obtained on the PlantVillage dataset varied from 85.53% (in case of *AlexNet::TrainingFromScratch::GrayScale::80-20*) to 99.34%(in case of *GoogLeNet::TransferLearning::Color::80-20*), hence showing strong promise of the deep learning approach for similar prediction problems. Table 1 shows the mean $F_1$ score, mean precision, mean recall, and overall accuracy across all our experimental configurations. All the experimental configurations run for a total of 30 epochs each, and they almost consistently converge after the first step down in the learning rate.

To address the issue of over-fitting, we vary the test set to train set ratio and observe that even in the extreme case of training on only 20% of the data and testing the trained model on the rest 80% of the data, the model achieves an overall accuracy of 98.21% (mean $F_1$-Score of 0.9820) in the case of *GoogLeNet::TransferLearning::Color::20-80*. As expected, the overall performance of both AlexNet and GoogLeNet do degrade if we keep increasing the test set to train set ratio (see Figure 4(d)), but the decrease in performance is not as drastic as we would expect if the model was indeed over-fitting. Figure 4(c) also shows that there is no divergence between the validation loss and the training loss, confirming that over-fitting is not a contributor to the results we obtain across all our experiments.

Among the AlexNet and GoogLeNet architectures, GoogLeNet consistently performs better than AlexNet 4(a), and based on the method of training, transfer learning always yields better results 4(b), both of which were expected.

The three versions of the dataset (color, gray-scale and segmented) show a characteristic variation in performance across all the experiments when we keep the rest of the experimental configuration constant. The models perform the best in case of the colored version of the dataset. When designing the experiments, we were concerned that the neural networks might only learn to pick up the inherent biases associated with the lighting conditions, the method and apparatus of

April 15, 2016 | 2

collection of the data. We therefore experimented with the gray-scaled version of the same dataset to test the model's adaptability in the absence of color information, and its ability to learn higher level structural patterns typical to particular crops and diseases. As expected, the performance did decrease when compared to the experiments on the colored version of the dataset, but even in the case of the worst performance, the observed mean $F_1$ score was 0.8524 (overall accuracy of 85.53%). The segmented versions of the whole dataset was also prepared to investigate the role of the background of the images in overall performance, and as shown in Figure 4(e), the performance of the model using segmented images is consistently better than that of the model using gray-scaled images, but slightly lower than that of the model using the colored version of the images.

Finally, while these approaches yield excellent results on the PlantVillage dataset which was collected in a controlled environment, we also assessed the model's performance on images sampled from trusted online sources such as academic agriculture extension services. Such images are not available in large numbers, and using a combination of automated download from Bing Image Search with a visual verification step by one of us (MS), we obtained a small, verified dataset of 121 images (see Supplementary Material for a detailed description of the process). By using the model trained using *GoogLeNet:Segmented:TransferLearning:80-20*, we obtained an overall accuracy of 31.40% in successfully predicting the correct class label (i.e. crop and disease information) from among 38 possible class labels. We note that a random classifier will obtain an average accuracy of only 2.63%. When providing the information about the crop that the particular image belongs to, the accuracy increases to 47.93%. Across all images, the correct class was in the top-5 predictions in 52.89% of the cases.

## Discussion

The performance of convolutional neural networks in object recognition and image classification has made tremendous progress in the past few years. [13] [14] [15] [16] [17]. Previously, the traditional approach for image classification tasks has been based on hand-engineered features such as SIFT[18], HoG[19], SURF[20], etc., and then to use some form of learning algorithm in these feature spaces. This led to the performance of all these approaches depending heavily on the underlying predefined features. Feature engineering itself is a complex and tedious process which needed to be revisited every time the problem at hand or the associated dataset changed considerably. This problem has occurred in all traditional attempts to detect plant diseases using computer vision as they leaned heavily on hand-engineered features, image enhancement techniques, and a host of other complex and labour-intensive methodologies. A few years ago, AlexNet[13] showed for the first time that end-to-end supervised training using a deep convolutional neural network architecture is a practical possibility even for image classification problems with a very large number of classes, beating the traditional approaches using hand-engineered features by a substantial margin in standard benchmarks. The absence of the labor-intensive phase of feature engineering and the generalizability of the solution makes them a very promising candidate for a practical and scaleable approach for computational inference of plant diseases.

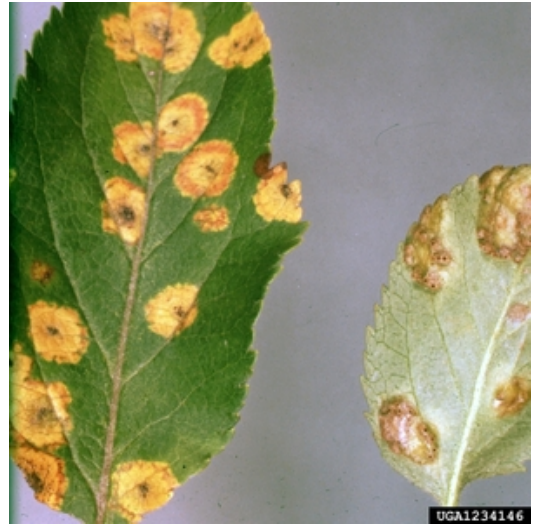

(a) Example image of a leaf suffering from Apple Cedar Rust, selected from the top-20 images returned by Bing Image search for the keywords "Apple Cedar Rust Leaves" on April 4th, 2016. Image Reference : *Clemson University - USDA Cooperative Extension Slide Series, Bugwood.org*

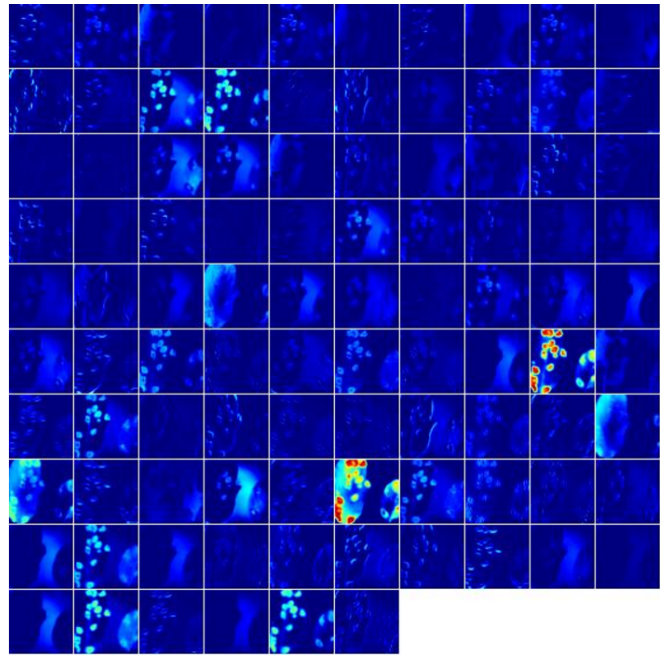

(b) Visualization of activations in the first convolution layer(*conv1*) of an AlexNet architecture trained using **AlexNet:Color:TrainFromScratch:80-20** when doing a forward pass on the image in Figure 3(a)

**Fig. 3.** Visualization of activations in the initial layers of an AlexNet architecture demonstrating that the model has learnt to efficiently activate against the diseased spots on the example leaf

Using the deep convolutional neural network architecture, we trained a model on images of plant leaves with the goal of classifying both crop species and the presence and identity of disease on images that the model had not seen before. Within the PlantVillage data set of 54,306 images containing 38 classes of 14 crop species and 26 diseases (or absence thereof), this goal has been achieved as demonstrated by the top accuracy of 99.35%. Thus, without any feature engineering, the model correctly classifies crop and disease from 38 possible classes in 993 out of 1000 images. Importantly, while the training of the model takes a lot of time (multiple hours on a high performance GPU cluster computer), the classification itself is very fast (less than a second on a CPU), and can thus easily be implemented on a smartphone. This presents a clear path towards smartphone-assisted crop disease diagnosis on a massive global scale.

Table 1. Mean $F_1$ score across various experiment configurations at the end of 30 Epochs. Each cell in the table represents the Mean $F_1$ score$_{\{mean\ precision,\ mean\ recall,\ overall\ accuracy\}}$ for the corresponding experimental configuration.

|  |  | AlexNet | | GoogLeNet | |
|---|---|---|---|---|---|
|  |  | **Transfer learning** | **Training from scratch** | **Transfer learning** | **Training from scratch** |
| **Train: 20%, Test: 80%** | Color | 0.9736$_{\{0.9742,\ 0.9737,\ 0.9738\}}$ | 0.9118$_{\{0.9137,\ 0.9132,\ 0.9130\}}$ | **0.9820**$_{\{0.9824,\ 0.9821,\ 0.9821\}}$ | 0.9430$_{\{0.9440,\ 0.9431,\ 0.9429\}}$ |
|  | Grayscale | 0.9361$_{\{0.9368,\ 0.9369,\ 0.9371\}}$ | 0.8524$_{\{0.8539,\ 0.8555,\ 0.8553\}}$ | **0.9563**$_{\{0.9570,\ 0.9564,\ 0.9564\}}$ | 0.8828$_{\{0.8842,\ 0.8835,\ 0.8841\}}$ |
|  | Segmented | 0.9724$_{\{0.9727,\ 0.9727,\ 0.9726\}}$ | 0.8945$_{\{0.8956,\ 0.8963,\ 0.8969\}}$ | **0.9808**$_{\{0.9810,\ 0.9808,\ 0.9808\}}$ | 0.9377$_{\{0.9388,\ 0.9380,\ 0.9380\}}$ |
| **Train: 40%, Test: 60%** | Color | 0.9860$_{\{0.9861,\ 0.9861,\ 0.9860\}}$ | 0.9555$_{\{0.9557,\ 0.9558,\ 0.9558\}}$ | **0.9914**$_{\{0.9914,\ 0.9914,\ 0.9914\}}$ | 0.9729$_{\{0.9731,\ 0.9729,\ 0.9729\}}$ |
|  | Grayscale | 0.9584$_{\{0.9588,\ 0.9589,\ 0.9588\}}$ | 0.9088$_{\{0.9090,\ 0.9101,\ 0.9100\}}$ | **0.9714**$_{\{0.9717,\ 0.9716,\ 0.9716\}}$ | 0.9361$_{\{0.9364,\ 0.9363,\ 0.9364\}}$ |
|  | Segmented | 0.9812$_{\{0.9814,\ 0.9813,\ 0.9813\}}$ | 0.9404$_{\{0.9409,\ 0.9408,\ 0.9408\}}$ | **0.9896**$_{\{0.9896,\ 0.9896,\ 0.9898\}}$ | 0.9643$_{\{0.9647,\ 0.9642,\ 0.9642\}}$ |
| **Train: 50%, Test: 50%** | Color | 0.9896$_{\{0.9897,\ 0.9896,\ 0.9897\}}$ | 0.9644$_{\{0.9647,\ 0.9647,\ 0.9647\}}$ | **0.9916**$_{\{0.9916,\ 0.9916,\ 0.9916\}}$ | 0.9772$_{\{0.9774,\ 0.9773,\ 0.9773\}}$ |
|  | Grayscale | 0.9661$_{\{0.9663,\ 0.9663,\ 0.9663\}}$ | 0.9312$_{\{0.9315,\ 0.9318,\ 0.9319\}}$ | **0.9788**$_{\{0.9789,\ 0.9788,\ 0.9788\}}$ | 0.9507$_{\{0.9510,\ 0.9507,\ 0.9509\}}$ |
|  | Segmented | 0.9867$_{\{0.9868,\ 0.9868,\ 0.9869\}}$ | 0.9551$_{\{0.9552,\ 0.9555,\ 0.9556\}}$ | **0.9909**$_{\{0.9910,\ 0.9910,\ 0.9910\}}$ | 0.9720$_{\{0.9721,\ 0.9721,\ 0.9722\}}$ |
| **Train: 60%, Test: 40%** | Color | 0.9907$_{\{0.9908,\ 0.9908,\ 0.9907\}}$ | 0.9724$_{\{0.9725,\ 0.9725,\ 0.9725\}}$ | **0.9924**$_{\{0.9924,\ 0.9924,\ 0.9924\}}$ | 0.9824$_{\{0.9825,\ 0.9824,\ 0.9824\}}$ |
|  | Grayscale | 0.9686$_{\{0.9689,\ 0.9688,\ 0.9688\}}$ | 0.9388$_{\{0.9396,\ 0.9395,\ 0.9391\}}$ | **0.9785**$_{\{0.9789,\ 0.9786,\ 0.9787\}}$ | 0.9547$_{\{0.9554,\ 0.9548,\ 0.9551\}}$ |
|  | Segmented | 0.9855$_{\{0.9856,\ 0.9856,\ 0.9856\}}$ | 0.9595$_{\{0.9597,\ 0.9597,\ 0.9596\}}$ | **0.9905**$_{\{0.9906,\ 0.9906,\ 0.9906\}}$ | 0.9740$_{\{0.9743,\ 0.9740,\ 0.9745\}}$ |
| **Train: 80%, Test: 20%** | Color | 0.9927$_{\{0.9928,\ 0.9927,\ 0.9928\}}$ | 0.9782$_{\{0.9786,\ 0.9782,\ 0.9782\}}$ | **0.9934**$_{\{0.9935,\ 0.9935,\ 0.9935\}}$ | 0.9836$_{\{0.9839,\ 0.9837,\ 0.9837\}}$ |
|  | Grayscale | 0.9726$_{\{0.9728,\ 0.9727,\ 0.9725\}}$ | 0.9449$_{\{0.9451,\ 0.9454,\ 0.9452\}}$ | **0.9800**$_{\{0.9804,\ 0.9801,\ 0.9798\}}$ | 0.9621$_{\{0.9624,\ 0.9621,\ 0.9621\}}$ |
|  | Segmented | 0.9891$_{\{0.9893,\ 0.9891,\ 0.9892\}}$ | 0.9722$_{\{0.9725,\ 0.9724,\ 0.9723\}}$ | **0.9925**$_{\{0.9925,\ 0.9925,\ 0.9924\}}$ | 0.9824$_{\{0.9827,\ 0.9824,\ 0.9822\}}$ |

However, there are a number of limitations at the current stage that need to be addressed in future work. First, when tested on a set of images taken under conditions different from the images used for training, the model's accuracy is reduced substantially, to 31.4%. It's important to note that this accuracy is much higher than the one based on random selection of 38 classes (2.6%), but nevertheless, a more diverse set of training data is needed to improve the accuracy. Our current results indicate that more (and more variable) data alone will be sufficient to substantially increase the accuracy, and corresponding data collection efforts are underway.

The second limitation is that we are currently constrained to the classification of single leaves, facing up, on a homogeneous background. While these are straightforward conditions, a real world application should be able to classify images of a disease as it presents itself directly on the plant. Indeed, many diseases don't present themselves on the upper side of leaves only (or at all), but on many different parts of the plant. Thus, new image collection efforts should try to obtain images from many different perspectives, and ideally from settings that are as realistic as possible.

At the same time, by using 38 classes that contain both crop species and disease status, we have made the challenge harder than ultimately necessary from a practical perspective, as growers are expected to know which crops they are growing. Given the very high accuracy on the PlantVillage dataset, limiting the classification challenge to the disease status won't have a measurable effect. However, on the real world data set, we can measure noticeable improvements in accuracy. To do this, we limit ourselves to crops where we have at least n>=2 or n>=3 classes per crop (to avoid trivial classification). In the n>=2 case, the dataset contains 33 classes distributed among 9 crops. Random guessing in such a dataset would achieve an accuracy of 0.273, while our model has an accuracy of 0.478. In the n>=3 case, the dataset contains 25 classes distributed among 5 crops. Random guessing in such a dataset would achieve an accuracy of 0.2, while our model has an accuracy of 0.411.

## Methods

**Dataset Description.** We analyze 54,306 images of plant leaves, which have a spread of 38 class labels assigned to them. Each class label is a crop-disease pair, and we make an attempt to predict the crop-disease pair given just the image of the plant leaf. Figure 1 shows one example each from every crop-disease pair from the PlantVillage dataset. In all the approaches described in this paper, we resize the images to 256 x 256 pixels, and we perform both the model optimization and predictions on these downscaled images.

Across all our experiments, we use three different versions of the whole PlantVillage dataset. We start with the PlantVillage dataset as it is, in color; then we experiment with a grayscaled version of the PlantVillage dataset, and finally we run all the experiments on a version of the PlantVillage dataset where the leaves were segmented, hence removing all the

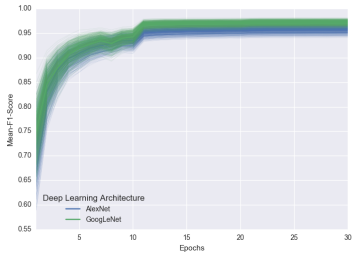 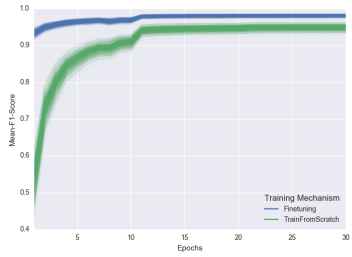 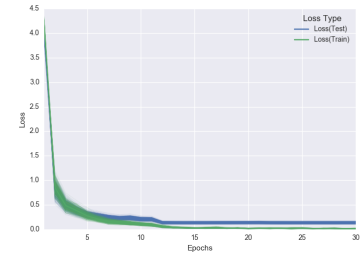

(a) Comparison of progression of mean $F_1$ score across all experiments, grouped by deep learning architecture

(b) Comparison of progression of mean $F_1$ score across all experiments, grouped by training mechanism

(c) Comparison of progression of train-loss and test-loss across all experiments.

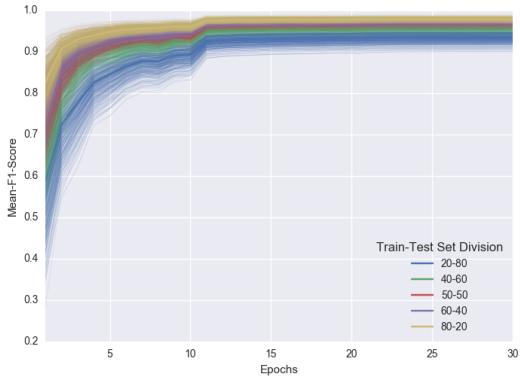 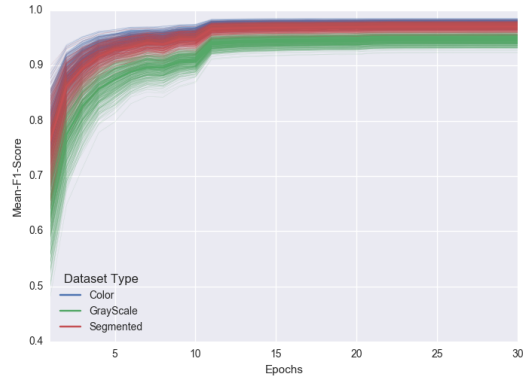

(d) Comparison of progression of mean $F_1$ score across all experiments, grouped by train-test set splits

(e) Comparison of progression of mean $F_1$ score across all experiments, grouped by dataset type

**Fig. 4.** Progression of mean $F_1$ score and loss through the training period of 30 epochs across all experiments, grouped by experimental configuration parameters. The intensity of a particular class at any point is proportional to the corresponding uncertainty across all experiments with the particular configurations. A similar plot of all the direct observations can be found in the Supplementary Material.

extra background information which might have the potential to introduce some inherent bias in the dataset due to the regularized process of data collection in case of PlantVillage dataset. Segmentation was automated by the means of a script tuned to perform well on our particular dataset. We chose a technique based on a set of masks generated by analysis of the color, lightness and saturation components of different parts of the images in several color spaces (Lab and HSB). One of the steps of that processing also allowed us to easily fix color casts, which happened to be very strong in some of the image collection subsets, thus removing another potential bias.

This set of experiments was designed to understand if the neural network actually learns the "notion" of plant diseases, or if it is just learning the inherent biases in the dataset. Figure 2 shows the different versions of the same leaf for a randomly selected set of leaves.

**Measurement of Performance.** To get a sense of how our approaches will perform on new unseen data, and also to keep a track of if any of our approaches are overfitting, we run all our experiments across a whole range of train-test set splits, namely 80-20 ( 80% of the whole dataset used for training, and 20% for testing), 60-40 ( 60% of the whole dataset used for training, and 40% for testing), 50-50 ( 50% of the whole dataset used for training, and 50% for testing), 40-60 ( 40% of the whole dataset used for training, and 60% for testing) and finally 20-80 ( 20% of the whole dataset used for training, and 80% for testing). It must be noted that in many cases, the PlantVillage dataset has multiple images of the same leaf (taken from different orientations), and we have the mappings of such cases for 41,112 images out of the 54,306 images; and during all these test-train splits, we make sure all the images of the same leaf goes either in the training set or the testing set. Further, for every experiment, we compute the mean precision, mean recall, mean $F_1$ score, along with the overall accuracy over the whole period of training at regular intervals (at the end of every epoch). We use the final mean $F_1$ score for the comparison of results across all of the different experimental configurations.

**Approach.** We evaluate the applicability of deep convolutional neural networks for the said classification problem. We focus on two popular architectures, namely AlexNet[13] and GoogLeNet[16], which were designed in the context of the "Large Scale Visual Recognition Challenge" (ILSVRC)[11] for the ImageNet dataset[12].

The AlexNet architecture follows the same design pattern as the LeNet-5[21] architecture from the 1990s. The LeNet-5 architecture variants are usually a set of stacked convolution layers followed by one or more fully connected layers. The convolution layers optionally may have a normalization layer and a pooling layer right after them, and all the layers in the network usually have ReLu non linear activation units associated with them. AlexNet consists of 5 convolution layers, followed by 3 fully connected layers, and finally ending with a softMax layer. The first two convolution layers (conv{1,2}) are each followed by a normalization and a pooling layer, and the last convolution layer(conv5) is followed by a single pooling layer. The final fully connected layer (fc8) has 38 outputs in our adapted version of AlexNet (equaling the total number of classes in our dataset), which feeds the softMax layer. All of the first 7 layers of AlexNet have a ReLu non-linearity activation unit associated with them, and the first two fully connected layers (fc{6,7}) have a dropout layer associated with them, with a dropout ratio of 0.5.

The GoogleNet architecture on the other hand is a much deeper and wider architecture with 22 layers, while still having considerably lower number of parameters ( 5 million parameters) in the network than AlexNet ( 60 million parameters). An application of the "network in network" architecture[22] in the form of the inception modules is a key feature of the GoogleNet architecture. The inception module uses parallel 1x1, 3x3 and 5x5 convolutions along with a max-pooling layer in parallel, hence enabling it to capture a variety of features in parallel. In terms of practicality of the implementation, the amount of associated computation needs to be kept in check, so they add 1x1 convolutions before the above mentioned 3x3, 5x5 convolutions (and also after the max-pooling layer) for dimensionality reduction. Finally, a filter concatenation layer simply concatenates the outputs of all these parallel layers. While this forms a single inception module, a total of 9 inception modules is used in the version of the GoogLeNet architecture that we use in our experiments. A more detailed overview of this architecture can be found for reference in [16].

We analyze the performance of both these architectures on the PlantVillage dataset by training the model from scratch in one case, and then by adapting already trained models (trained on the ImageNet dataset) using transfer learning. In case of transfer learning, we do not limit the learning of the rest of the layers, and we instead just reset the weights of layer fc8 in case AlexNet; in case of GoogLeNet, we similarly do not limit the learning of the rest of the layers but instead just reset the weights of the loss{1,2,3}/classifier layers.

To summarize, we have a total of 60 experimental configurations, which vary on the following parameters :

1. Choice of deep learning architecture
   - AlexNet
   - GoogLeNet

2. Choice of training mechanism
   - Transfer Learning
   - Training from Scratch

3. Choice of dataset type
   - Color
   - Gray scale
   - Leaf Segmented

4. Choice of training-testing set distribution
   - Train: 80% , Test: 20%
   - Train: 60% , Test: 40%
   - Train: 50% , Test: 50%
   - Train: 40% , Test: 60%
   - Train: 20% , Test: 80%

Throughout this paper, we have used the notation of *Architecture:TrainingMechanism:DatasetType:Train-Test-Set-Distribution* to refer to particular experiments. For instance, to refer to the experiment using the GoogLeNet architecture, which was trained using transfer learning on the gray-scaled PlantVillage dataset on a train-test set distribution of 60-40, we will use the notation *GoogLeNet:TransferLearning:GrayScale:60-40*.

Each of these 60 experiments runs for a total of 30 epochs, where one epoch is defined as the number of training iterations in which the particular neural network has completed a full pass of the whole training set. The choice of 30 epochs was made based on the empirical observation that in all of these experiments, the learning always converged well within 30 epochs (as is evident from the aggregated plots (Figure 2) across all the experiments).

To enable a fair comparison between the results of all the experimental configurations, we also tried to standardize the hyperparameters across all the experiments, and we used the following hyperparameters in all of the experiments :

- Solver type: Stochastic Gradient Descent
- Base learning rate: 0.005
- Learning rate policy: Step (decreases by a factor of 10 every 30/3 epochs)
- Momentum: 0.9
- Weight decay: 0.0005
- Gamma: 0.1
- Batch size: 24 (in case of GoogLeNet), 100 (in case of AlexNet)

All the above experiments were conducted using our own fork of Caffe[1] [23], which is a fast, open source framework for deep learning. The basic results such as the overall accuracy can also be replicated using a standard instance of caffe.

**ACKNOWLEDGMENTS.** We thank Boris Conforty for help with the segmentation. We thank Kelsee Baranowski, Ryan Bringenberg and Megan Wilkerson for taking the images and Kelsee Baranowski for image curation. We thank Anna Sostarecz, Kaity Gonzalez, Ashtyn Goodreau, Kalley Veit, Ethan Keller, Parand Jalili, Emma Volk, Nooeree Samdani, Kelsey Pryze for additional help with image curation. We thank EPFL, and the Huck Institutes at Penn State University for support. We are particularly grateful for access to EPFL GPU cluster computing resources.

---

[1] https://github.com/salathegroup/caffe